\RequirePackage{amsmath}
\documentclass{article}

\usepackage[T1]{fontenc}
\usepackage{graphicx}
\usepackage{algorithm}
\usepackage{algpseudocode}
\usepackage{dsfont}
\usepackage{multirow}
\usepackage[caption=false]{subfig}
\usepackage{amsmath}
\usepackage{hyperref}

\usepackage{arxiv}

\usepackage[utf8]{inputenc} 
\usepackage[T1]{fontenc}    
\usepackage{hyperref}       
\usepackage{url}            
\usepackage{booktabs}       
\usepackage{amsfonts}       
\usepackage{nicefrac}       
\usepackage{microtype}      
\usepackage{lipsum}
\usepackage{graphicx}

\title{Personalized Student Knowledge Modeling for Future Learning Resource Prediction}

\author{
    \textbf{Soroush Hashemifar} \\
    Department of Computer Science\\
    University at Albany - SUNY\\
    Albany, NY 12222, USA\\
    \texttt{shashemifar@albany.edu}\\
    \and 
    \textbf{Sherry Sahebi\thanks{Corresponding author}} \\
    Department of Computer Science\\
    University at Albany - SUNY\\
    Albany, NY 12222, USA\\
    \texttt{ssahebi@albany.edu}\\
}

\begin{document}
\maketitle
\begin{abstract}
Despite advances in deep learning for education, student knowledge tracing and behavior modeling face persistent challenges:
limited personalization, inadequate modeling of diverse learning activities (especially non-assessed materials), and overlooking the interplay between knowledge acquisition and behavioral patterns.
Practical limitations, such as fixed-size sequence segmentation, frequently lead to the loss of contextual information vital for personalized learning.
Moreover, reliance on student performance on assessed materials limits the modeling scope, excluding non-assessed interactions like lectures.
To overcome these shortcomings, we propose Knowledge Modeling and Material Prediction (KMaP), a stateful multi-task approach designed for personalized and simultaneous modeling of student knowledge and behavior. 
KMaP employs clustering-based student profiling to create personalized student representations, improving predictions of future learning resource preferences.
Extensive experiments on two real-world datasets confirm significant behavioral differences across student clusters and validate the efficacy of the KMaP model.
\end{abstract}


\section{Introduction}\label{sec:introduction}
Online education provides exceptional flexibility, enabling learning anytime and anywhere, which attracts many learners. 
Research indicates that online platforms can enhance educational experiences by recommending and personalizing learning resources~\cite{Tapalova:PersonalAIEd,Abdelshiheed:DRLtutor,Huang:flippedclassroom}.
Two essential elements of this personalized approach are Knowledge Tracing (KT)~\cite{Corbett:BKT} and Behavior Modeling (BM)~\cite{Shen:learning,Yang:implicit}. 
KT models the progression of students' knowledge over time~\cite{Dowling:automata} and BM yields insights into student behaviors, including their engagement levels with various learning materials~\cite{Goldberg:engagement} and future material selection~\cite{Pandey:moocrec,Wu:exercise}. 
As prior research shows that student behavior and knowledge acquisition are linked~\cite{Dong:engagement}, exploring simultaneous modeling of student behavior and knowledge could provide new insights in both areas.
Furthermore, such simultaneous modeling could benefit pedagogy by including both student knowledge and personal preferences in personalized instruction and adaptive scaffolding.
For example, these models could facilitate addressing the students' knowledge gaps while preserving learner autonomy. 

Few recent studies have explored simultaneous modeling of student behavior and knowledge, intending to improve both tasks~\cite{Zhao:KTBM,Zhao:Pareto-TAMKOT}. 
However, further research is essential to enhance the adaptability of these models to learn from diverse learning materials and introduce personalization to them.
While some recent works model learning from different types of learning materials~\cite{Abdi:open,Zhao:GMKT}, many KT models still focus on the assessed types (like problems and quizzes).
Additionally, recent models, especially the deep ones, often segment learning activities into fixed-length subsequences, disregarding long-term sequential interdependencies. 
This process disrupts the continuity of knowledge and behavior representations, making it more difficult to trace individual learning paths and failing to account for the unique differences among students.
Common solutions, such as learning student-specific parameters or extended segments, often lead to overparameterization, overfitting, and vanishing gradient issues, which hinder the ability to achieve effective personalization.

To address the above challenges, this paper presents Knowledge Modeling and Material Prediction (KMaP), a stateful multi-task approach addressing KT personalization and prediction of student's preferences for future materials.
KMaP dynamically updates student representations by clustering-based profiling to ensure state consistency while enabling personalization across learning trajectories.
Using contrastive training, it learns to identify contextually relevant resources among various material types. 
Experiments demonstrate that personalization reveals latent student behavioral groups, and KMaP surpasses existing methods in modeling student preferences for future learning materials.
KMaP's source code is available at \href{https://github.com/persai-lab/2025-contrastiveMultiTaskKTBM}{https://github.com/persai-lab/2025-AIED-KMaP}.

\section{Related Work}\label{sec:relatedwork}
\noindent\textbf{Knowledge Tracing.} 
\sloppy Knowledge tracing focuses on understanding how students acquire knowledge by interacting with learning materials. 
The foundational Bayesian Knowledge Tracing (BKT)~\cite{Corbett:BKT} employs a Markov process to deduce latent knowledge states from student performance in assessed learning materials, like problems. 
Later, various advanced models have been proposed based on BKT.
For example, Dynamic Bayesian Knowledge Tracing (DBKT)~\cite{Kaser:DBKT} enhances BKT by incorporating relationships between knowledge concepts (KCs) via dynamic Bayesian networks. 
Progress in deep learning spearheaded models like Deep Knowledge Tracing (DKT)~\cite{Piech:DKT} that utilize recurrent neural networks (RNNs) to analyze learning interactions. 
Further, attention-based methods like AKT~\cite{Pu:AKT} and SAINT~\cite{Choi:SAINT} model knowledge decay and intricate question-answer dynamics and memory-augmented neural networks (MANN)~\cite{Santoro:MANN}, exemplified by DKVMN~\cite{Zhang:DKVMN}, track student understanding in fine-grained concept features. 
Other recent models, like TAMKOT~\cite{Zhao:TAMKOT} and GMKT~\cite{Zhao:GMKT} have been proposed in recent years to address novel challenges like multi-type KT that captures the knowledge transfer between assessed and non-assessed material types.
To the best of our knowledge, no existing deep knowledge tracing models currently accommodate personalization for diverse types of learning resources.

\noindent\textbf{Behavior Modeling.} 
Behavior modeling examines student interactions with educational materials to identify learning patterns and preferences. 
Student behavior can have multiple aspects, like procrastination~\cite{Zhao:curbproc}, engagement~\cite{Goldberg:engagement}, and choice behavior~\cite{He:leveraging}.
Recent models aim to simultaneously consider behavior and knowledge, with behavior indicating activity timings in many cases.  
For example, LFKT~\cite{Khajah:integrating} addresses simultaneous student behavior and knowledge modeling challenges by considering problem difficulties. 
MBFKT~\cite{Diao:MBFKT} integrates behavioral features, such as interaction time intervals and repeated study of the same knowledge components (KCs), into knowledge tracing models. 
Pareto-TAMKOT~\cite{Zhao:Pareto-TAMKOT} advances this by analyzing transitions between different types of materials, utilizing a modified LSTM network to forecast upcoming material types and student performance. 
KTBM~\cite{Zhao:KTBM} further enhances this by incorporating a memory-augmented LSTM to dynamically update student knowledge states and predict future performance and material types. 
However, no studies have addressed the challenge of modeling behavior by predicting the next learning materials in a student's learning journey, particularly in contexts that involve both assessed and non-assessed learning resources.

\section{Knowledge Modeling and Material Prediction (KMaP)}\label{sec:KMaP}

\subsection{Problem Formulation}\label{subsec:formulation}
We aim to propose a personalized model that captures individual student learning and behavior patterns simultaneously, while identifying student similarities.
Consider a system with a set of students ($S$) that interact with assessed ($Q$, e.g., questions) and non-assessed ($E$, e.g., lectures) learning materials. We model each learning activity as a tuple $\langle s, i_t, r_t, z_t \rangle$, where $i_t$ specifies the learning material that student $s$ has interacted with, $r_t$ represents the student's performance or score in $i_t$, and $z_t \in \{0, 1\}$ indicates the material type (0 for assessed and 1 for non-assessed). 
For non-assessed materials, $r_t$ is set to a padding value.  
Given a student's interaction history, our objectives are to predict (1) the type of the student's next activity ($z_{t+1}$), (2) the learning material that the student will interact with ($i_{t+1}$), and (3) the student's performance on the upcoming assessed material ($r_{t+1}$), provided $z_{t+1} = 0$. 

\subsection{The KMaP Model}\label{subsec:model}
\begin{figure}
    \includegraphics[width=\textwidth]{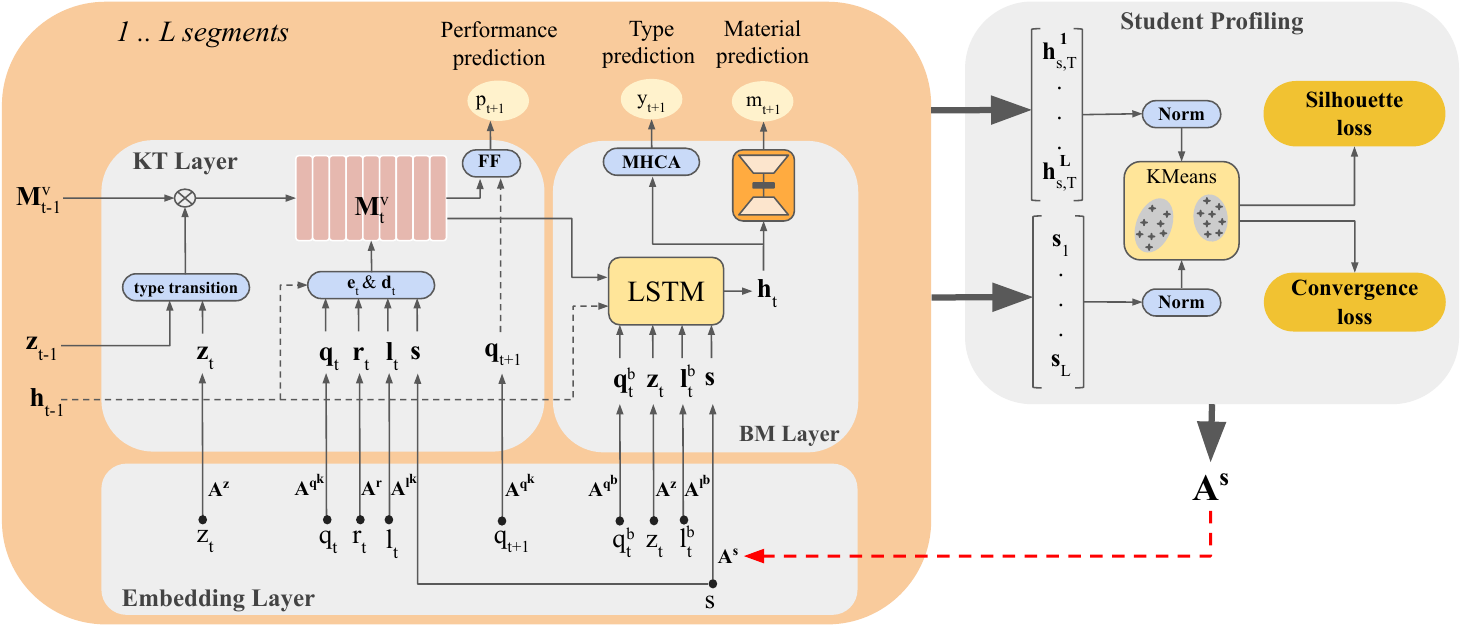}
    \caption{The proposed model architecture for KMaP.}
    \label{fig:archKMaP}
\end{figure}

KMaP consists of four components: Embedding layer, Knowledge Tracing (KT), Behavior Modeling (BM), and clustering-based student profiling. An overview of  KMaP's architecture is illustrated in Fig.~\ref{fig:archKMaP}.

\subsection{Embedding Layer}\label{subsec:embedding}
This layer transforms each learning activity $\langle s, i_t, r_t, z_t \rangle$ into input features suitable for further processing.
Students are projected to embeddings ($\boldsymbol{s}$) in a latent space using the transformation matrix $\boldsymbol{A}^s \in \mathcal{R}^{|S| \times d_s}$.
Similarly, questions ($q_t$), lectures ($l_t$), responses ($r_t$), and material types ($z_t$) are transformed into respective embeddings ($\boldsymbol{q}^k_t$, $\boldsymbol{l}^k_t$, $\boldsymbol{r}_t$, $\boldsymbol{z}_t$) using matrices $\boldsymbol{A}^{q^k} \in \mathcal{R}^{|Q| \times d_{q^k}}$, $\boldsymbol{A}^{l^k} \in \mathcal{R}^{|E| \times d_{l^k}}$, $\boldsymbol{A}^r \in \mathcal{R}^{2 \times d_r}$, and $\boldsymbol{A}^z \in \mathcal{R}^{2 \times d_z}$.
For behavior modeling, another set of material embeddings ($\boldsymbol{q}^b_t, \boldsymbol{l}^b_t$) are generated using projection matrices $\boldsymbol{A}^{q^b} \in \mathcal{R}^{|Q| \times d_{q^b}}$ and $\boldsymbol{A}^{l^b} \in \mathcal{R}^{|E| \times d_{l^b}}$, providing flexibility to learn specific representations for BM.

Representing $N_c$ latent knowledge concepts~\cite{Zhang:DKVMN}, each learning material corresponds to a vector $\boldsymbol{w}_t \in \mathcal{R}^{N_c}$, which serves as attentional weights in MANN. 
MANN learns static KC features ($\boldsymbol{M}^k_t$) to compute $\small \boldsymbol{w}_t = softmax(\boldsymbol{q}_t \cdot \boldsymbol{M}^k_t)$.
However, KMaP prevents this computationally intensive calculation by shared, pre-trained $\boldsymbol{w}_t$ parameters.
It projects materials into a latent space using transformation matrices $\boldsymbol{A}^w_q \in \mathcal{R}^{|Q| \times N_c}$ and $\boldsymbol{A}^w_l \in \mathcal{R}^{|E| \times N_c}$, and subsequently applies \textit{softmax} to the resulting weights for normalization.

\subsection{Knowledge Tracing}\label{subsec:knowledge}
We organize the KT component into two key parts: knowledge monitoring and student performance prediction.

 \noindent\textbf{Knowledge Monitoring.} 
KMaP tracks student $s$'s knowledge progression across $N_c$ concepts using a dynamic value matrix $\boldsymbol{M}^{v,i}_{s,t} \in \mathcal{R}^{N_c \times d_k}$, with $d_k$ denoting feature dimension.
KMaP updates this matrix by an \textit{erase-followed-by-add} mechanism, which personalizes behavior-based forgetting and knowledge acquisition based on the student's profile $\boldsymbol{s}$ and previous behavior state $\boldsymbol{h}^i_{s,t-1}$, as specified in equations~\ref{eq:erasevec}-\ref{eq:knowstate}.
\small  \begin{equation}  
    \label{eq:erasevec}
    \boldsymbol{e}_t = \sigma((1 - z_t) {\boldsymbol{E}_q}^T [\boldsymbol{q}_t \oplus \boldsymbol{r}_t] + z_t {\boldsymbol{E}_l}^T \boldsymbol{l}_t + {\boldsymbol{E}_h}^T \boldsymbol{h}^i_{s,t-1} + {\boldsymbol{E}_s}^T \boldsymbol{s} + \boldsymbol{b}_e)
\end{equation} \normalsize
\small  \begin{equation}  
    \label{eq:addvec}
    \boldsymbol{d}_t = Tanh((1 - z_t) {\boldsymbol{D}_q}^T [\boldsymbol{q}_t \oplus \boldsymbol{r}_t] + z_t {\boldsymbol{D}_l}^T \boldsymbol{l}_t + {\boldsymbol{D}_b}^T \boldsymbol{h}^i_{s,t-1} + {\boldsymbol{D}_s}^T \boldsymbol{s} + \boldsymbol{b}_d)
\end{equation} \normalsize
\small  \begin{equation}  
    \label{eq:knowstate}
    \boldsymbol{M}^{v,i}_{s,t}(i) = [Tanh(\boldsymbol{W}_{tr} [\boldsymbol{z}_{t-1} \oplus \boldsymbol{z}_t]) \cdot \boldsymbol{M}^{v,i}_{s,t-1}](i) \times [1 - \boldsymbol{w}_t(i) \boldsymbol{e}_t] + \boldsymbol{w}_t(i) \boldsymbol{d}_t
\end{equation} \normalsize

$\boldsymbol{E}_*$ and $\boldsymbol{D}_*$ weight matrices transform all entities to the same dimensions.
$\boldsymbol{b}_e$ and $\boldsymbol{b}_d$ are biases.
$\boldsymbol{W}_{tr} \in \mathcal{R}^{d_v \times 2d_z}$ manages transitions between assessed and non-assessed materials across consecutive time-steps.

\noindent\textbf{Student Performance Prediction.} 
To estimate student $s$'s performance on the next given question, $\boldsymbol{q}_{t+1}$, we calculate student mastery, $\boldsymbol{c}_{t+1}$, as: 
\small  \begin{equation}  
    \label{eq:readcontent}
    \boldsymbol{c}_{t+1} = \sum^{N_c}_{i=1} w_{t+1}(i)[Tanh(\boldsymbol{W}_{tr}[\boldsymbol{z}_{t} \oplus \boldsymbol{z}_q]) \cdot \boldsymbol{M}^{v,i}_{s,t}](i)
\end{equation} \normalsize

The question type embedding $\boldsymbol{z}_q$ ($\boldsymbol{z}_q = [1 \ 0] \cdot \boldsymbol{A}^z$) selects the first row of $\boldsymbol{A}^z$, which represents question-type features. 
Subsequently, the response correctness probability $p_{t+1}$ is estimated by integrating combined predictive features $\boldsymbol{c}_{t+1}$, question embedding $\boldsymbol{q}_{t+1}$, and behavior state $\boldsymbol{h}^i_{s,t}$ through a feed-forward network with \textit{sigmoid} activation function.

\subsection{Behavior Modeling}\label{subsec:behavior}
The behavior modeling component consists of three key elements: behavior monitoring, next material prediction, and next material type prediction.

\noindent\textbf{Behavior Monitoring.}
Similar to KTBM~\cite{Zhao:KTBM}, a modified LSTM tracks student behavior, however, we personalize this using student profile $\boldsymbol{s}$, as shown in equation~\ref{eq:lstmcell}, with $\boldsymbol{h}^i_{s,t}$ and $\boldsymbol{m}^i_{s,t}$ denoting hidden and cell states.
\small  \begin{equation}  
    \label{eq:lstmcell}
    \boldsymbol{h}^i_{s,t}, \boldsymbol{m}^i_{s,t} = \text{LSTMCell}(z_t, \boldsymbol{q}^b_t, \boldsymbol{l}^b_t, \boldsymbol{z}_t, \boldsymbol{h}^i_{s,t-1}, \boldsymbol{M}^{v,i}_{s,t}, \boldsymbol{s}) 
\end{equation} \normalsize

\begin{figure}[t]
    \includegraphics[width=\textwidth]{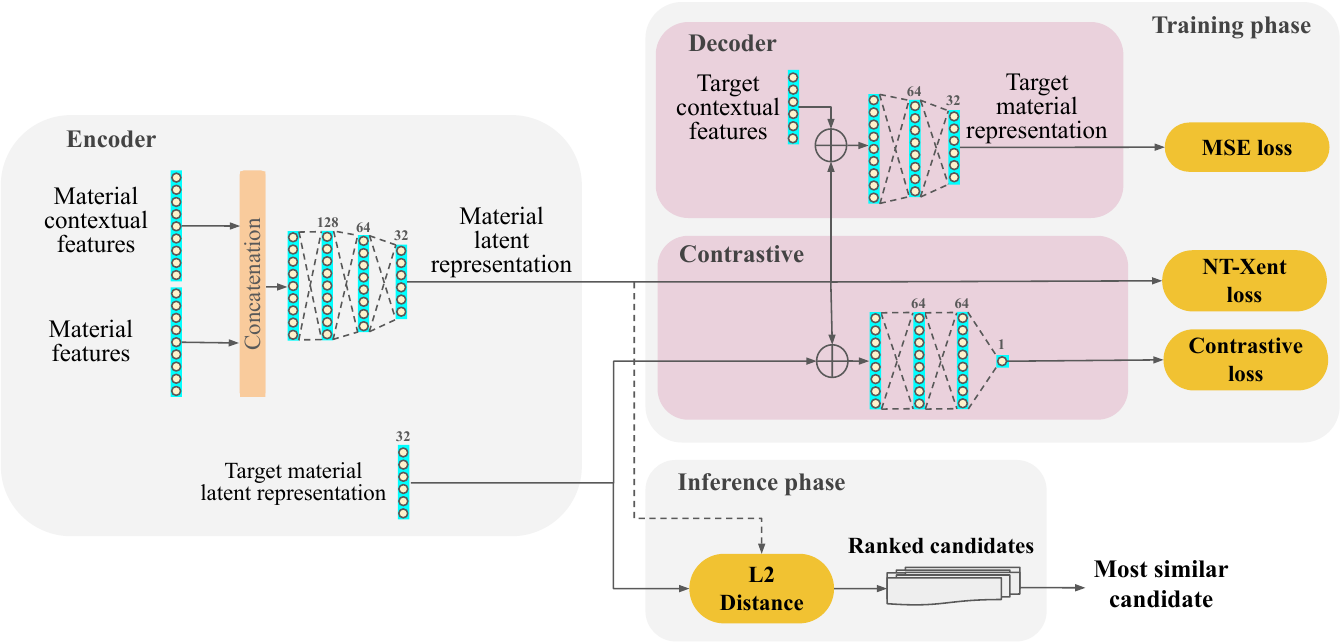}
    \caption{Proposed encoding method for learning material prediction in KMaP.}
    \label{fig:encoderKMaP}
\end{figure}

\noindent\textbf{Next Material prediction.}
We utilize an encoder-decoder architecture to predict the next material (Fig.~\ref{fig:encoderKMaP}). 
The encoder maps the current material and its contextual features into a latent space. 
During training, the contrastive approach pairs the next material ($\boldsymbol{q}^b_{t+1}$ or $\boldsymbol{l}^b_{t+1}$) with $k$ negative random samples to learn latent features that align closely with the next material while remaining distant from the negative candidates. 
During inference, KMaP takes $k+1$ latent features of $k$ negative samples and one ground-truth learning material and retrieves the material that closely matches its prediction.
The following equations illustrate the encoder's output for current assessed material ($\boldsymbol{a}_{t,q} \in \mathcal{R}^{d_v}$) and next assessed material ($\boldsymbol{p}_{t,q} \in \mathcal{R}^{d_v}$) projected into $d_v$-dimensional latent space:
\small  \begin{equation}  
    \label{eq:qencanchor}
\begin{aligned}
    \boldsymbol{a}_{t,q} = FF^q_{enc}([\boldsymbol{h_t} \oplus \boldsymbol{q}^b_t \oplus & \boldsymbol{w}_{q_t} \oplus \boldsymbol{c}_{q_t} \oplus \boldsymbol{w}_{q_{t-1}} \oplus \boldsymbol{c}_{q_{t-1}} \oplus \boldsymbol{w}_{l_{t-1}} \oplus \boldsymbol{c}_{l_{t-1}}])
\end{aligned}
\end{equation} \normalsize
\small  \begin{equation}  
    \label{eq:qencpos}
\begin{aligned}
    \boldsymbol{p}_{t,q} = FF^q_{enc}([\boldsymbol{h_t} \oplus \boldsymbol{q}^b_{t+1} \oplus & \boldsymbol{w}_{q_{t+1}} \oplus \boldsymbol{c}_{q_{t+1}} \oplus 
    \boldsymbol{w}_{q_t} \oplus \boldsymbol{c}_{q_t} \oplus \boldsymbol{w}_{l_t} \oplus \boldsymbol{c}_{l_t}])
\end{aligned}
\end{equation} \normalsize

The equations for non-assessed materials ($\boldsymbol{a}_{t,l}$ and $\boldsymbol{p}_{t,l}$) are symmetrical to the assessed ones, with $\boldsymbol{z}_q$ replaced by $\boldsymbol{z}_l$ (via $[0  \  1] \cdot \boldsymbol{A}^z$) in calculating knowledge mastery $\boldsymbol{c}_{t}$ (Eq.~\ref{eq:readcontent}).
Unlike the inference phase, KMaP is informed about the positive (next) and negative (random) materials during training, enabling the model to differentiate them. 
However, during inference, KMaP ranks all candidate materials, according to the $L_2$ distance between $\boldsymbol{a}_{t,.}$ and the encoded features of the candidate.
Eventually, it selects the candidate with the lowest distance.

The decoder reconstructs the next material representation by integrating both current material information and the surrounding contextual features. 
It works based on the assumption that there exists an intrinsic relationship between materials and contextual features associated with them.
For assessed materials, equations~\ref{eq:qdecanchor} and~\ref{eq:qdecpos} formulate the decoder. Similar equations are used for the non-assessed ones. 
Notably, the decoder is used exclusively during the training phase and is inaccessible during the evaluation and inference stages.
\small  \begin{equation}  
    \label{eq:qdecanchor}
\begin{aligned}
    \boldsymbol{\hat{q}}_t = FF^q_{dec}([\boldsymbol{p}_{t,q} \oplus \boldsymbol{w}_{q_t} & \oplus \boldsymbol{c}_{q_t} \oplus \boldsymbol{l}^b_t \oplus \boldsymbol{w}_{l_t} \oplus \boldsymbol{c}_{l_t} \oplus \boldsymbol{q}^b_{t+1} \oplus \boldsymbol{w}_{q_{t+1}} \\
    & \oplus \boldsymbol{c}_{q_{t+1}} \oplus \boldsymbol{l}^b_{t+1} \oplus \boldsymbol{w}_{l_{t+1}} \oplus \boldsymbol{c}_{l_{t+1}}])
\end{aligned}
\end{equation} \normalsize
\small  \begin{equation}  
    \label{eq:qdecpos}
\begin{aligned}
    \boldsymbol{\hat{q}}_{t+1} = FF^q_{dec}([\boldsymbol{a}_{t,q} \oplus & \boldsymbol{w}_{q_{t+1}} \oplus \boldsymbol{c}_{q_{t+1}} \oplus \boldsymbol{l}^b_{t+1} \oplus \boldsymbol{w}_{l_{t+1}} \oplus \boldsymbol{c}_{l_{t+1}} \\
    & \oplus \boldsymbol{q}^b_t \oplus \boldsymbol{w}_{q_t} \oplus \boldsymbol{c}_{q_t} \oplus \boldsymbol{l}^b_t \oplus \boldsymbol{w}_{l_t} \oplus \boldsymbol{c}_{l_t}])
\end{aligned}
\end{equation} \normalsize

\noindent\textbf{Next Material Type prediction.}
To predict the upcoming material type, KMaP employs a modified Multi-Head Cross-Attention (MHCA) mechanism, with the student profile as the query and contextual features as key and value.
These contextual features include question embedding $\boldsymbol{q}^b_t$, student response $\boldsymbol{r}_t$, lecture representation $\boldsymbol{l}^b_t$, and behavioral hidden states $\boldsymbol{h}^i_{s,t}$.
MHCA is followed by a linear projection layer with \textit{sigmoid} activation function for predicting $y_{t+1}$. 

\subsection{Personalized Student Profiling}\label{subsec:profiling}
KMaP achieves student state personalization and continuity through two main strategies: a stateful structure and a cluster-based student profiling component.

\noindent\textbf{Stateful Structure.}
KMaP processes long sequences by retaining intermediate student states across segmented learning histories of size $T$ via a stateful network.
This ensures continuous tracking of progress, concept mastery, and behavioral patterns.
Instead of resetting hidden states per segment, KMaP incrementally updates knowledge states, behavioral hidden states, and cell states.
It initializes these states with prior terminal states for each segment $i \in \{1 .. L\}$:
\small  \begin{equation}  
    \label{eq:stateinit}
    \boldsymbol{M}^{v,i}_{s,0} = \begin{cases}
        \small \mathcal{N}(\mu, \sigma) & \text{if first batch} \\
        \small \boldsymbol{M}^{v,i-1}_{s,T} & \text{otherwise}
    \end{cases}, \quad
    \boldsymbol{h}^i_{s,0} = \begin{cases}
        \small \boldsymbol{0} & \text{if first batch} \\
        \small \boldsymbol{h}^{i-1}_{s,T} & \text{otherwise}
    \end{cases}
\end{equation} \normalsize

The formulation for the initial cell state $\boldsymbol{m}^i_{s,0}$ is similar to hidden state $\boldsymbol{h}^i_{s,0}$.

\noindent\textbf{Student Cluster-based Profiling.}
To identify student behavioral patterns and improve the model generalizability, KMaP employs a three-step clustering approach.
It starts by deriving student behaviors as hidden states $\{\boldsymbol{h}^i_{s,T} | i : 1.. L\}$ per learning segment.
Students are then clustered using the KMeans algorithm based on the average of their corresponding hidden states across all segments.
Finally, cluster centroids $c_a$ are computed. 

During training, KMaP updates student embeddings within each segment through backpropagation of segment loss, without changing the initial student embeddings. 
Hence, $L$ distinct segment-specific student embeddings $\mathcal{I}_s=\{\boldsymbol{s}_{i} | i : 1.. L\}$ are derived at the end of this process. 
While learning distinct embeddings would lead to model flexibility, we expect these embeddings to be similar to each other.
So, as the second step, we incorporate an additional convergence loss function that ensures the minimum distance between each student's embeddings:
\small  \begin{equation}    
    \label{eq:lconv}
\mathcal{L}_{conv} = \frac{1}{|S|} \sum_{s \in S} \sum_{(\boldsymbol{s}_{i},\boldsymbol{s}_{j}) \in \mathcal{I}_s} \lVert \boldsymbol{s}_{i} - \boldsymbol{s}_{j} \rVert_2
\end{equation} \normalsize

Also, we compute generalized representative embeddings $\bar{V}_s$ by normalizing and averaging segment-specific embeddings $\boldsymbol{s}_{i}$.
Expecting $\bar{V}_s$ to be aligned with overall student behavior, we incorporate a silhouette loss~\cite{Rousseeuw:silhouettes} that enforces proximity to cluster centroids, using intra and inter-cluster distances $d_{ic}$ and $d_{nc}$, where $\tau$ is for temperature scaling of the fraction to be between -1 and +1:
\small  \begin{equation}  
    \label{eq:lsilhouette}
    \mathcal{L}_{silhouette} = \frac{1}{|S|} \bigg[\sum_{s \in S} \exp{(\frac{-(d_{nc} - d_{ic}) / \max(d_{ic}, d_{nc})}{\tau})} \bigg]
\end{equation} \normalsize

\subsection{Optimization}\label{subsec:optimization}
We propose a two-stage optimization comprising: 1) an inner loop addressing student response prediction, upcoming learning material selection, and material type classification, and 2) an outer loop dedicated to student clustering.

\noindent\textbf{Inner loop.}
To train the autoencoder we optimize a Mean Squared Error loss $\mathcal{L}_{rec}$ between the reconstructed and the ground-truth material representations.
\small  \begin{equation}  
    \label{eq:lrec}
    \mathcal{L}_{rec} = MSE(\boldsymbol{q}^b_t, \boldsymbol{\hat{q}}_t) + MSE(\boldsymbol{q}^b_{t+1}, \boldsymbol{\hat{q}}_{t+1}) + MSE(\boldsymbol{l}^b_t, \boldsymbol{\hat{l}}_t) + MSE(\boldsymbol{l}^b_{t+1}, \boldsymbol{\hat{l}}_{t+1})
\end{equation} \normalsize

We formulate next-material prediction contrastively, differentiating positive (chosen) from negative (non-chosen) samples.
A hybrid loss integrates a Binary Cross-Entropy (BCE)-based contrastive loss ($\mathcal{L}_{cont}$) with our proposed variant of Normalized Temperature-Scaled Cross-Entropy (NT-Xent) to help with numerous negative samples per each positive sample.
\small  \begin{equation}  
    \label{eq:lcon}
    \mathcal{L}_{cont} = - (\sum \boldsymbol{y}_{t,q} \log \boldsymbol{\hat{y}}_{t,q} + \sum \boldsymbol{y}_{t,l} \log \boldsymbol{\hat{y}}_{t,l})
\end{equation} \normalsize
Contrastive probabilities ($\boldsymbol{\hat{y}}_{t,q}, \boldsymbol{\hat{y}}_{t,l}$) are computed by feeding anchor ($\boldsymbol{a}_{t,.}$), positive ($\boldsymbol{p}_{t,.}$), and negative ($\boldsymbol{n}^k_{t,.}$) material representations, through a feed-forward network, followed by \textit{softmax}, evaluated by ground-truth labels ($\boldsymbol{y}_{t,q}, \boldsymbol{y}_{t,l}$).
The NT-Xent loss ensures latent proximity between the anchor (current material) and positive material.
However, the original NT-Xent loss function~\cite{Oord:cpc} produces negative values when the anchor-positive distance ($\small \lVert \boldsymbol{a}_{t,l} - \boldsymbol{p}_{t,l} \rVert_2$) is significantly smaller than the anchor-negative distance ($\small \sum_k \lVert \boldsymbol{a}_{t,q} - \boldsymbol{n}^k_{t,q} \rVert_2$). 
To mitigate this, we reformulate NT-Xent by exponentiating the loss, guaranteeing non-negativity: 
\small  \begin{equation}  
    \label{eq:lntxent}
    \mathcal{L}_{ntxent} = \frac{\sum_{k} \exp({-\lVert \boldsymbol{a}_{t,q} - \boldsymbol{n}^k_{t,q} \rVert_2})}{\exp({-\lVert \boldsymbol{a}_{t,q} - \boldsymbol{p}_{t,q} \rVert_2})} + 
    \frac{\sum_{k} \exp({-\lVert \boldsymbol{a}_{t,l} - \boldsymbol{n}^k_{t,l} \rVert_2})}{\exp({-\lVert \boldsymbol{a}_{t,l} - \boldsymbol{p}_{t,l} \rVert_2})}
\end{equation} \normalsize

For student performance and material type predictions, we use BCE losses $\mathcal{L}_{\text{perf}}$ and $\mathcal{L}_\text{type}$ across all time-steps. 
The final optimization function is: 
\small  \begin{equation}  
    \label{eq:allLoss}
    \min \enspace \frac{1}{T} \sum_{t = 1}^T (\mathcal{L}_{cont} + \mathcal{L}_{rec} + \mathcal{L}_{ntxent}  + \mathcal{L}_{\text{perf}} + \mathcal{L}_\text{type})
\end{equation} \normalsize

\noindent\textbf{Outer loop.}
The outer-loop optimization focuses on the clustering and refinement of student embeddings, based on the losses presented in Section~\ref{subsec:profiling}.
Algorithm~\ref{alg:studentalgo} outlines our two-stage optimization: the inner loop employs SGD to obtain segment-specific behavioral states and embeddings, resetting gradients for parameter updates (lines 4-12).
The outer loop refines student embeddings by averaging normalized embeddings and behavioral states (lines 13-17). 
KMeans clusters students by behavioral states (line 18), computes centroids (line 19), and updates student embeddings, given the clustering objectives (line 20).

\begin{algorithm}[t]
  \caption{KMaP Student Profiling Algorithm}
    \label{alg:studentalgo}
\begin{algorithmic}[1]
    \State {\bfseries Input:} Student embeddings $\boldsymbol{s}$, Number of epochs $E$, Number of clusters $nClusters$
    \For{$epoch = 1$ to $E$}
        \State Initialize empty containers $V = \{\;\}$ (segment-specific student embeddings) and $B = \{\;\}$ (segment-specific behavioral states)
        \For{$iteration = 1$ to $N_{\text{batches}}$}
            \State Perform feed-forward pass
            \State Compute loss gradient $\nabla{\ell}$ using Eq.~\ref{eq:allLoss}
            \State Update student embeddings: $\boldsymbol{s}_{\text{new}} \gets \boldsymbol{s} - \text{lr} \times \nabla_{\boldsymbol{s}}{\ell}$
            \State Store updated embeddings: $V \gets V \cup \{\boldsymbol{s}_{\text{new}}\}$
            \State Extract behavioral hidden states $s_h$ and store: $B \gets B \cup \{\boldsymbol{s}_h\}$
            \State Reset gradient: $\nabla_{\boldsymbol{s}}{\ell} \gets 0$
            \State Backpropagate $\nabla{\ell}$
        \EndFor
        \State Initialize empty containers $\bar{V} = \{\;\}$ and $\bar{B} = \{\;\}$ for average normalized embeddings
        \For{each student $s$}
            \State Normalize embeddings:
            \[
            V_s \gets \frac{V_s}{\lVert V_s \rVert_2}, \quad B_s \gets \frac{B_s}{\lVert B_s \rVert_2}
            \]
            \State Compute mean embeddings for student $s$:
            \[
            \bar{V}_s \gets \bar{V}_s \cup \left\{\frac{1}{|\mathcal{I}_{\mathit{s}}|} \sum_{i \in \mathcal{I}_{\mathit{s}}} V_s[i]\right\}, \quad 
            \bar{B}_s \gets \bar{B}_s \cup \left\{\frac{1}{|\mathcal{I}_{\mathit{s}}|} \sum_{i \in \mathcal{I}_{\mathit{s}}} B_s[i]\right\}
            \]
        \EndFor
        \State Perform clustering: $c_a \gets \text{KMeans}(\bar{B}, nClusters)$
        \State Compute centroids and their respective embeddings ($c_n$ and $c_m$) using normalized student embeddings: $c_n, c_m \gets \text{NearestCentroid}(\bar{V}, c_a)$
        \State Backpropagate clustering loss gradient $\nabla_{\boldsymbol{s}}{\ell_{\text{Silhouette}}+\ell_{\text{conv}}}$ 
    \EndFor
\end{algorithmic}
\end{algorithm}

\section{Experiments}\label{sec:experiments}

\subsection{Data}\label{subsec:data}
We assess KMaP using two real-world datasets: EdNet and Junyi.
\textbf{EdNet}~\cite{Choi:EdNet} is provided by Santa~\cite{Riiid:Santa}, a system that structures questions into mandatory sequential bundles but allows autonomous learner navigation among other resources, such as selecting lectures or purchasing courses. 
It consists of four levels, with our analysis focusing on the third level from 1,000 students who interacted with both assessed materials (questions) and non-assessed learning resources (lectures). 
\textbf{Junyi}~\cite{CMU:JunyiAcademy} is a Chinese mathematics e-learning platform that features exercises with prerequisite relationships but permits flexible navigation within unlocked topics. 
Learners can progress from foundational topics to more advanced ones. 
The platform offers assessed problems, and non-assessed hints, which provide supplementary assistance in solving the problems.

\subsection{Experimental Setup}\label{subsec:setup}
KMaP is evaluated on predicting students' next learning material choice, next material type, and response.
Two ablation studies are conducted: (i) KMaP-P, which retains personalization exclusively; and (ii) KMaP-M, which retains the material prediction head exclusively.
KMaP is implemented in PyTorch, optimized via Adam, with hyperparameters specified in Table~\ref{table:hyperparams}.
The contrastive strategy incorporates 5 (training) and 99 (evaluation/inference) randomly sampled negative candidates. 
Data preprocessing involves segmenting activities into 100-length padded sequences (80\% train, 20\% test). 
Original sequence order is preserved to maintain dependencies, while user shuffling is used to enhance diversity. 
For generalization, during training, 10\% of materials are randomly dropped for 50\% of users per batch and replaced with padding tokens. 
\begin{table}[t]
    \centering
    \renewcommand{\arraystretch}{1.2}
    \caption{Learned best hyperparameters.}
    \label{table:hyperparams}
    \begin{tabular}{c|ccccccccc|cccc}
        \hline
        Dataset & $d_s$ & $d_q$ & $d_r$ & $d_l$ & $d_z$ & $d_{q^b}$ & $d_{l^b}$ & $N_c$ & $d_v$ & lr & nClusters & cutoff & $\tau$ \\
        \hline
        EdNet & 32 & 64 & 32 & 32 & 32 & 32 & 32 & 8 & 32 & 0.1 & 3 & 5 & 0.1\\
        Junyi & 32 & 32 & 32 & 32 & 32 & 32 & 32 & 8 & 32 & 0.01 & 3 & 5 & 0.1\\
        \hline
    \end{tabular}
\end{table} 

\subsection{Material Prediction Comparison}\label{subsec:matcomp}
By comparing against LSTM~\cite{Hochreiter:LSTM}, SASRec~\cite{Kang:SASRec}, and TE-DCN~\cite{Li:TE-DCN}, Tables~\ref{table:assessedmat} and~\ref{table:nonassessedmat} demonstrate the superior performance of KMaP and KMaP-M for both material types in EdNet, as well as for assessed materials in Junyi.
More specifically, KMaP matches SASRec on Junyi's non-assessed materials, whereas KMaP-M surpasses all baselines.
This emphasizes the limited personalization benefits in Junyi's structured environment, which restricts behavioral variability due to its prerequisite-guided sequencing.
Conversely, EdNet's diverse, open-ended learning pathways enhance personalization efficacy, supported by richer student interactions. 
The notable prediction gap for EdNet's non-assessed materials can be attributed to its design: 
AI-recommended questions versus student-selected lectures, with the lecture's contextual alignment to prior questions, enabling KMaP's effective utilization of conceptual cues (Eq.~\ref{eq:qencanchor}–\ref{eq:qdecpos}).

\begin{table}[t]
    \centering
    \renewcommand{\arraystretch}{1.2}
    \caption{Assessed learning material prediction. $*$ indicates significance ($p\leq0.01$)}
    \label{table:assessedmat}
    \begin{tabular}{ccccccc}
        \hline
        \multirow{2}{*}{Methods} & \multicolumn{3}{c}{EdNet} & \multicolumn{3}{c}{Junyi} \\
        \cline{2-7}
        & HR@5 & NDCG@5 & MRR & HR@5 & NDCG@5 & MRR \\
        \hline
        LSTM            & 0.0574* & 0.0350* & 0.0579* & 0.2976* & 0.2244* & 0.2253* \\
        SASRec          & 0.2582* & 0.2384* & 0.2562* & 0.7009* & 0.6660* & 0.6689* \\
        TE-DCN          & 0.3150* & 0.2763* & 0.2905* & 0.6807* & 0.6442* & 0.6472* \\
        \hline
        KMaP-M      & $\underline{0.4210}$ & $\underline{0.3515}$ & $\underline{0.3604}$ & \textbf{0.8838} & \textbf{0.8368} & \textbf{0.8293} \\
        KMaP        & \textbf{0.4220} & \textbf{0.3523} & \textbf{0.3608} & $\underline{0.8730}$ & $\underline{0.8247}$ & $\underline{0.8181}$ \\
        \hline
    \end{tabular}
\end{table} 
\begin{table}
    \centering
    \renewcommand{\arraystretch}{1.2}
    \caption{Non-assessed learning material prediction. $*$ indicates significance ($p\leq0.01$)}
    \label{table:nonassessedmat}
    \begin{tabular}{ccccccc}
        \hline
        \multirow{2}{*}{Methods} & \multicolumn{3}{c}{EdNet} & \multicolumn{3}{c}{Junyi} \\
        \cline{2-7}
        & HR@5 & NDCG@5 & MRR & HR@5 & NDCG@5 & MRR \\
        \hline
        LSTM            & 0.0550* & 0.0320* & 0.0548* & 0.2015* & 0.1349* & 0.1417* \\
        SASRec          & 0.1144* & 0.0889* & 0.1103* & $\underline{0.8074*}$ & $\underline{0.7743*}$ & $\underline{0.7750*}$ \\
        TE-DCN          & 0.2315* & 0.1663* & 0.1819* & 0.7273* & 0.6849* & 0.6861* \\
        \hline
        KMaP-M      & \textbf{0.8511} & \textbf{0.7649} & \textbf{0.7455} & \textbf{0.8391} & \textbf{0.7994} & \textbf{0.7948} \\
        KMaP        & $\underline{0.8289}$ & $\underline{0.7332}$ & $\underline{0.7123}$ & $\underline{0.7987}$ & $\underline{0.7540}$ & $\underline{0.7502}$ \\
        \hline
    \end{tabular}
\end{table}

\subsection{Response Prediction Comparison}\label{subsec:respcomp}
Table~\ref{table:performance} demonstrates KMaP effectiveness on response (performance) prediction against DKVMN~\cite{Zhang:DKVMN}, SAINT~\cite{Choi:SAINT}, GMKT~\cite{Zhao:GMKT}, and KTBM~\cite{Zhao:KTBM}.
The results indicate KMaP variants outperform or match all baselines except KTBM, which uniquely employs multi-objective Pareto optimization to balance KT and material-type prediction.
This clarifies the importance of appropriately balancing task-specific losses to enhance model efficacy.
Ablation studies show KMaP-P's superior AUC on EdNet, highlighting personalization benefits in flexible environments, while KMaP-M excels in Junyi's structured curriculum, where personalization is less effective.
In particular, AUC discrepancies for DKVMN compared to the literature stem from our strategy of retained sequence order and data augmentation (Section~\ref{subsec:setup}).

\begin{table}[t]
    \centering
    \renewcommand{\arraystretch}{1.2}
    \caption{Student performance prediction. $*$ indicates significance ($p\leq0.01$)}
    \label{table:performance}
    \begin{tabular}{c|cccc|ccc}
        \hline
        Dataset & DKVMN & SAINT & GMKT & KTBM & KMaP-P & KMaP-M & KMaP \\
        \hline
        EdNet & 0.6069* & $\underline{0.6524}$ & 0.5520* & \textbf{0.6981} & 0.6519 & 0.6252 & 0.6370 \\
        Junyi & 0.6251* & 0.8088* & 0.8101* & \textbf{0.8864} & 0.8472 & $\underline{0.8585}$ & 0.8564 \\
        \hline
    \end{tabular}
\end{table}

\subsection{Type Prediction Comparison}\label{subsec:typecomp}
Comparing KMaP with GMKT~\cite{Zhao:GMKT}, KTBM~\cite{Zhao:KTBM}, and LSTM, the results (Table~\ref{table:type}) demonstrate predicting materials alongside student performance enhances material type prediction (KMaP vs. GMKT), but optimal task balance remains challenging (KMaP vs. KTBM).
Ablation studies reveal comparable performances of KMaP variants, which consistently outperform GMKT and LSTM models.
This suggests that neither material prediction nor personalization, when considered independently, significantly improves the performance of the base model on EdNet.
Although personalization provides slight benefits in certain cases, its overall impact remains limited, highlighting the challenge of isolating individual factors in this task.
\begin{table}
    \centering
    \renewcommand{\arraystretch}{1.2}
    \caption{Material type prediction. $*$ indicates significance ($p\leq0.01$)}
    \label{table:type}
    \begin{tabular}{c|ccc|ccc}
        \hline
        Dataset & LSTM & GMKT & KTBM & KMaP-P & KMaP-M & KMaP \\ 
        \hline
        EdNet & 0.8394* & 0.8442* & \textbf{0.9028} & $\underline{0.8871}$ & 0.8859 & 0.8836 \\
        Junyi & 0.8900* & 0.9008* & \textbf{0.9311} & 0.9176 & $\underline{0.9235}$ & 0.9171 \\
        \hline
    \end{tabular}
\end{table} 

\subsection{Students' Behavioral Analysis}\label{subsec:behanalysis}
\begin{figure}
    \centering
    \subfloat[]{\includegraphics[width=0.25\linewidth]{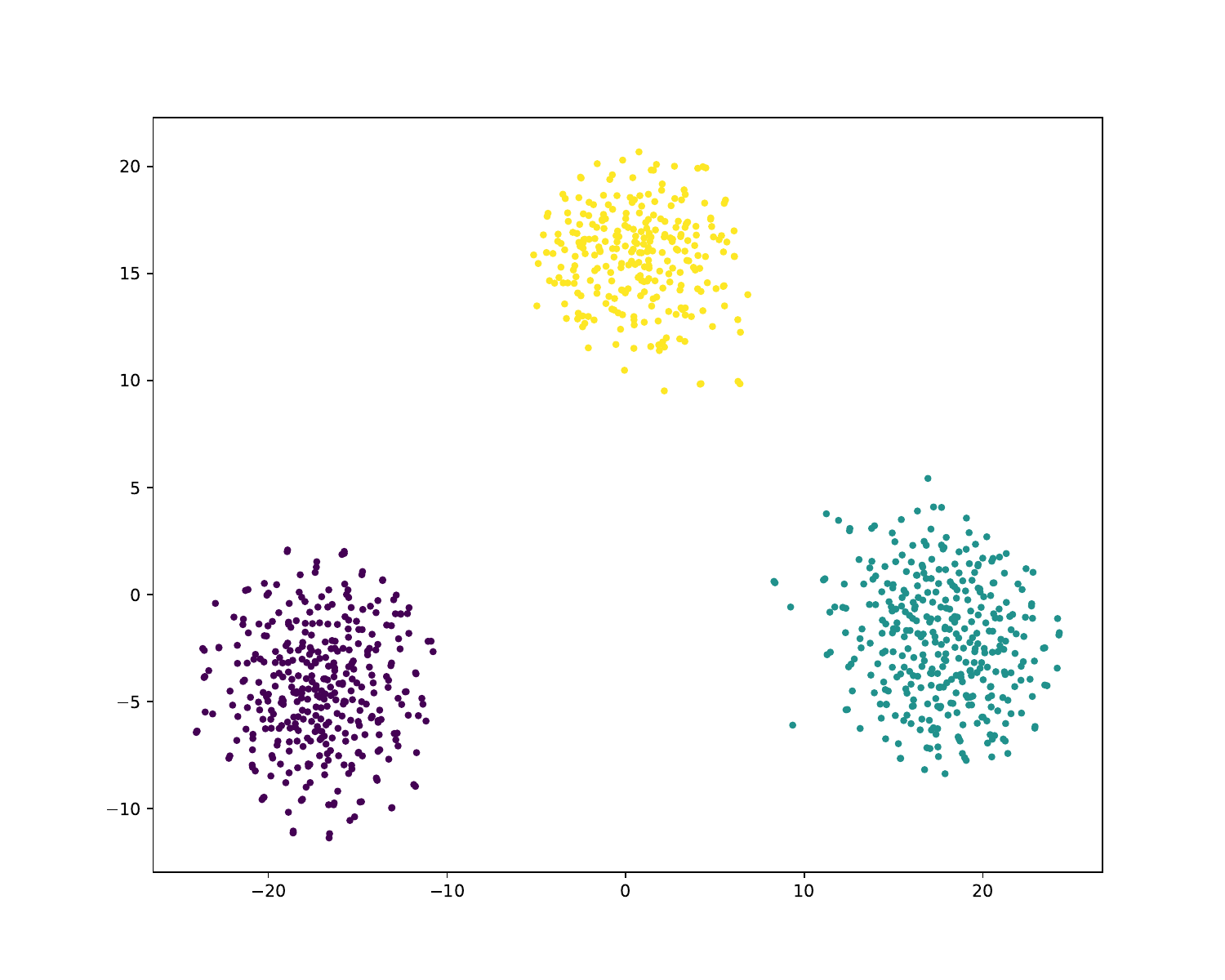}\label{fig:studentclustersA}}
    \subfloat[]{\includegraphics[width=0.25\linewidth]{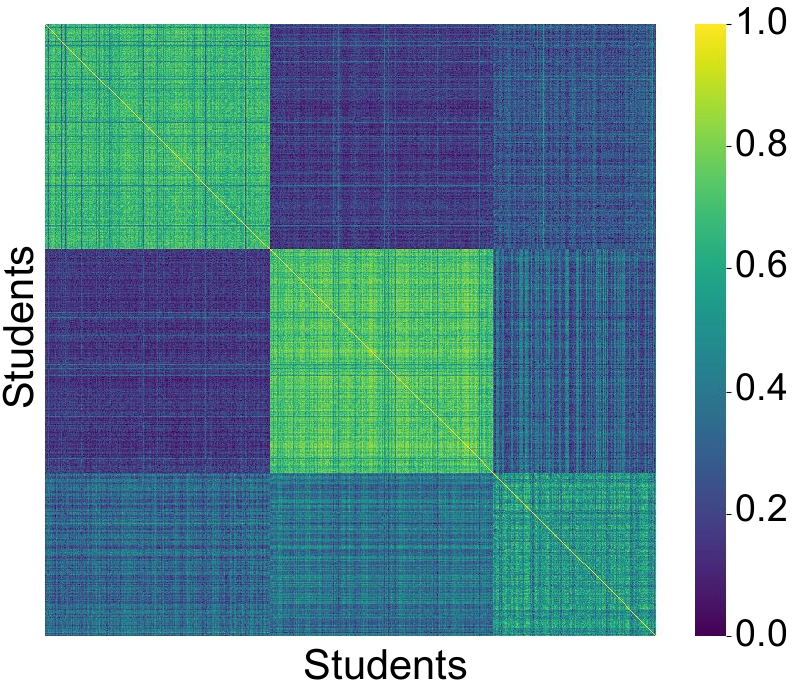}\label{fig:studentclustersB}}
    \subfloat[]
    {\includegraphics[width=0.25\linewidth]{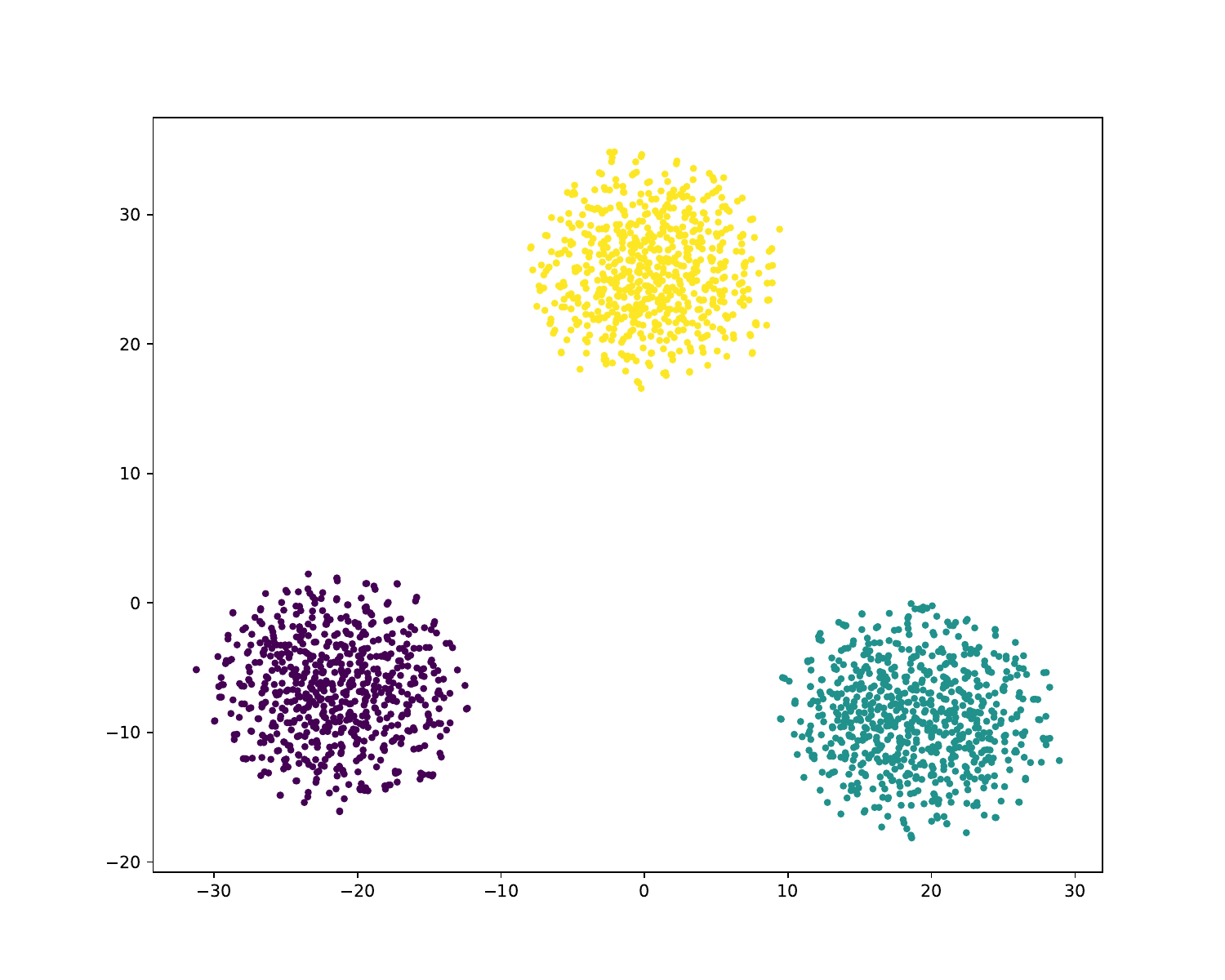}\label{fig:studentclustersC}}
    \subfloat[]{\includegraphics[width=0.25\linewidth]{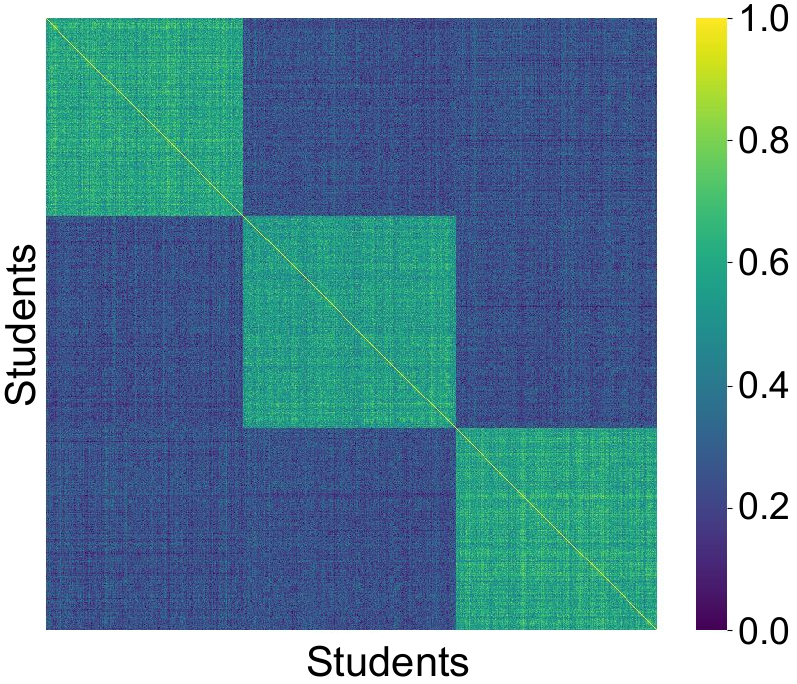}\label{fig:studentclustersD}}
    \caption{Visualization of student embeddings on EdNet (a \& b) and Junyi (c \& d)}
    \label{fig:studentclusters}
\end{figure}

\begin{figure}
    \centering
    \subfloat[]{\includegraphics[width=0.46\linewidth]{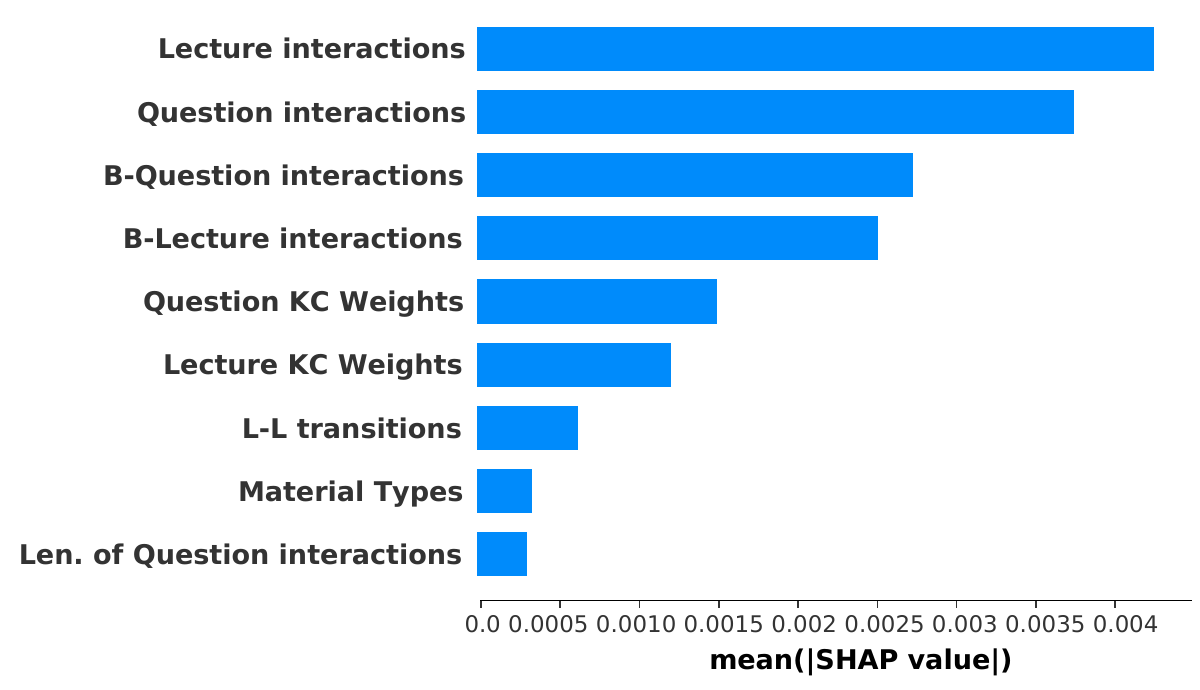}\label{fig:shapA}}
    \subfloat[]{\includegraphics[width=0.45\linewidth]{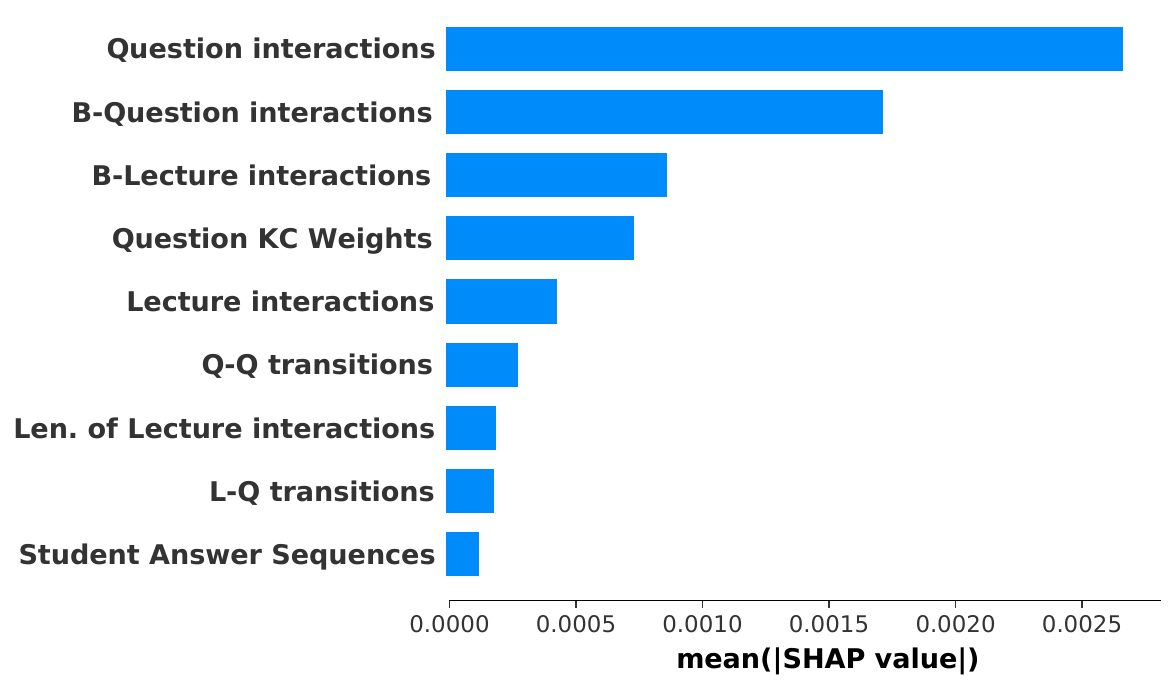}\label{fig:shapB}}
    \caption{SHAP plots of sample clusters of EdNet (a), and Junyi (b).}
    \label{fig:shaps}
\end{figure}

Fig.~\ref{fig:studentclusters} demonstrates KMaP's effective identification of distinct student groups through scatter plots and heatmaps.
EdNet clusters capture the behavioral tendencies of lecture-focused students, in contrast to problem-solving ones. 
However, Junyi's patterns distinguish learners who learn from mistakes from those who emphasize consolidation. 
These insights are derived using SHAP (SHapley Additive exPlanations)~\cite{Lundberg:shap} plots, shown in Fig.~\ref{fig:shaps}.
These plots illustrate the features that contribute most to a selected cluster, highlighting only those with non-zero contributions.
The selected cluster serves to facilitate a clear explanation of the key factors.
The features comprise sequences of (1) question and lecture embeddings (via $\boldsymbol{A}^{q^k}$ and $\boldsymbol{A}^{l^k}$) and their sequence lengths, (2) behavioral embeddings (via $\boldsymbol{A}^{q^b}$ and $\boldsymbol{A}^{l^b}$), (3) KC correlation weights, (4) material types and transition frequency, and (5) student responses to assessed materials.

The selected EdNet cluster is predominantly shaped by interactions with questions and lectures, as well as KCs associated with both materials.
Additionally, lecture transitions, types, and the length of interaction histories play a crucial role, with lectures having a particularly notable impact.
Conversely, the selected Junyi cluster is more sensitive to question interactions.
It formed clusters by distinctions related to KCs of questions, question transitions, lecture-to-question transitions, and student responses.
Similar patterns of behavior are observed, with minor variations, in other clusters as well.
These observations indicate dataset-specific learning behaviors and the underlying latent factors driving cluster formation.
Ultimately, the tightly bounded clusters in Fig.~\ref{fig:studentclusters} suggest a potential over-separation issue, which requires future research on inter- and intra-cluster variance analysis.

\section{Conclusions and Future Work}\label{sec:conclusions}
This study introduced Knowledge Modeling and Material Prediction (KMaP), a stateful multi-task approach designed to enhance learning resource prediction while personalizing knowledge tracing. 
KMaP models students' preferences as they interact with questions and lectures, aiming to predict student's response to the next question and the materials they are more likely to choose next.
Using contrastive learning and student profiling, KMaP effectively models knowledge progression and behavior. 
Experimental results showed 
higher effectiveness of personalization in the EdNet dataset compared to Junyi. 
Further, our analyses provided insight into student clusters and feature importance variations across datasets. 
In future research, we will explore advanced multi-objective optimization methods to effectively balance the objectives of our final loss function.

\begin{credits}
\noindent\textbf{\ackname} 
This paper is based on work partially supported by the National Science Foundation under Grant Number 2047500.

\noindent\textbf{\discintname}
The authors have no competing interests. 
\end{credits}

\bibliographystyle{unsrt}  
\bibliography{main}

\end{document}